\RequirePackage[utf8]{inputenc}
\RequirePackage[nointegrals]{wasysym}
\DeclareUnicodeCharacter{266B}{\twonotes}
\documentclass[11pt]{article}
\usepackage[preprint]{acl}

\usepackage{times}
\usepackage{latexsym}
\usepackage[T1]{fontenc}
\usepackage[utf8]{inputenc}
\usepackage{microtype}
\usepackage{inconsolata}
\usepackage{graphicx}
\usepackage{booktabs}
\usepackage{multirow}
\usepackage{amsmath,amssymb}
\usepackage{mathrsfs}
\usepackage{xcolor}
\usepackage{colortbl}

\title{BiCAA: Bidirectional Credit Assignment for Search-Augmented Agent}
\author{
      Yibin Huang\textsuperscript{1},
      Bin Xu\textsuperscript{1},
      Hailong Cao\textsuperscript{1},
      Conghui Zhu\textsuperscript{1}\thanks{Corresponding author.}
      \\
      \textsuperscript{1}Faculty of Computing, Harbin Institute of Technology
  }

\begin{document}
\maketitle
\begin{abstract}
By iteratively retrieving and integrating external evidence, multi-step search enables search-augmented agents to tackle complex question answering. However, vanilla GRPO assigns rewards based solely on final answers, providing limited guidance for intermediate search steps. To provide fine-grained process feedback, we evaluate the value of each search step: it should acquire evidence that advances problem solving while avoiding redundant retrieval. Building on these complementary criteria, we introduce BiCAA, which combines Forward Solvability Gain (FSG) and Subsequent Reasoning Gain (SRG) for bidirectional credit assignment. To reward progress in evidence acquisition, FSG quantifies changes in ground-truth answer likelihood after each interaction. SRG compares the subsequent action's likelihood with and without the current interaction to approximately measure its support for subsequent reasoning, using this support as a proxy signal for non-redundant search to encourage effective and efficient search. BiCAA calibrates local feedback through asymmetric fusion; for policy optimization, it integrates accumulated process rewards with terminal outcome supervision. Across seven open-domain QA benchmarks and two Qwen backbones, experiments show improvements in average F1, particularly on multi-hop tasks.
\end{abstract}

\section{Introduction}
Knowledge-intensive question answering~\cite{lewis2020retrieval} often requires iterative retrieval and evidence integration rather than a single search. Search-augmented agents gather external information, refine subsequent queries, and decide when sufficient evidence has been collected to answer~\cite{yao2022react,jiang2023active}. Reinforcement learning algorithms such as Group Relative Policy Optimization (GRPO) have been widely used to train these agents~\cite{shao2024deepseekmath,zeng2025reinforcing}. However, when rewards depend solely on final answers, GRPO provides limited guidance on the contributions of individual search steps. This sparse supervision motivates process-level credit assignment for multi-step search~\cite{lightman2024let}.

We argue that a valuable search step should acquire evidence that advances problem solving while avoiding redundant retrieval. Increases in answer likelihood can reflect progress toward solving the problem~\cite{wang2026information}, but repeated evidence may also improve this score. We therefore additionally use the current interaction's likelihood-based support for the next action as a proxy signal for non-redundant search, treating answer progress and subsequent support as complementary evaluation criteria.

Based on this insight, we propose BiCAA, a bidirectional credit assignment framework that combines Forward Solvability Gain (FSG) and Subsequent Reasoning Gain (SRG) to construct dense search-turn rewards. FSG compares the ground-truth answer's average token log-likelihood before and after each search interaction, providing an answer-directed signal for evidence acquisition. SRG compares the likelihood of the same observed next action with and without the current interaction to approximately assess its additional support for subsequent reasoning beyond earlier history. This comparison treats the current reasoning, query, and retrieved results as a complete search turn. We normalize the length-adjusted likelihood ratio within each trajectory to obtain relative weights for modulating FSG. Since repeated interactions may add little support beyond earlier history, this signal serves as a proxy for non-redundant search and complements FSG's answer-directed feedback.

BiCAA integrates the two signals through asymmetric modulation: FSG determines the direction of local answer-progress feedback, while SRG adjusts its strength according to relative subsequent support, reinforcing positive feedback and attenuating negative penalties for higher-support steps. BiCAA then combines process rewards with terminal outcome rewards under unified advantage estimation, retaining global supervision while assigning fine-grained credit to individual search steps.

Experiments on seven open-domain QA benchmarks spanning in-domain and out-of-domain tasks yield average F1 scores of 53.0 and 54.4 with Qwen2.5-7B-Instruct and Qwen3-8B, respectively, with particularly strong gains on multi-hop tasks. On Qwen3-8B, BiCAA exceeds the reported outcome-only GRPO average by 3.5 points. Ablations and behavioral analyses further support the effectiveness of its components and show reduced redundant search.

Our contributions are summarized as follows:
\begin{itemize}
  \item We propose a complementary perspective on search-step evaluation: a valuable interaction should acquire evidence that advances problem solving and support subsequent reasoning beyond earlier search efforts. Considering these criteria jointly connects evidence acquisition with effective, non-redundant search.
  \item We instantiate this perspective in BiCAA, a bidirectional credit assignment framework that jointly evaluates answer progress and subsequent support through asymmetric modulation and unified advantage estimation, providing dense feedback for search interactions.
  \item Experiments demonstrate that BiCAA reduces redundant search behavior and achieves competitive answer quality.
\end{itemize}

\section{Related Work}

\paragraph{RL Post-Training for Reasoning and Agents.}
RL post-training improves LLM reasoning through critic-based PPO~\cite{schulman2017proximal}, critic-free RLOO and REINFORCE++~\cite{ahmadian2024back,hu2025reinforce++}, and group-relative methods such as GRPO, DAPO, and Dr.GRPO~\cite{shao2024deepseekmath,yu2026dapo,liu2025understanding}.
Search-R1~\cite{jin2025search} trains interleaved reasoning and search with outcome rewards and retrieved-token masking.
ReSearch~\cite{chen2025research} learns when and how to search as part of the reasoning chain without supervised reasoning traces.
ZeroSearch~\cite{sun2025zerosearch} elicits search behavior through simulated retrieval and curriculum rollouts.
BiCAA retains group-based optimization but supplements trajectory-level outcome supervision with rewards for individual search steps.

\paragraph{Credit Assignment in Agentic Search.}
Process-level credit assignment targets intermediate decisions through reward design or finer-grained optimization.
GiGPO~\cite{feng2026group} refines group-relative credit by comparing actions under similar intermediate states.
Turn-level reward design~\cite{zeng2025reinforcing} provides localized feedback at each interaction rather than only at the final response.
POAD~\cite{wen2024reinforcing} decomposes language actions for finer-grained policy updates.
IGPO~\cite{wang2026information} addresses sparse outcome supervision using increases in the ground-truth answer's average token log-likelihood as turn-level rewards.
StepSearch~\cite{zheng2025stepsearch} combines reference-document coverage gains, document-repetition penalties, and query alignment with reference keywords to reward informative, non-redundant retrieval.
BiCAA builds on answer-likelihood gain by asymmetrically modulating it with likelihood-based subsequent support for complete search interactions, using this support as a proxy for non-redundant search without reference queries or supporting-document annotations.

\paragraph{Retrospective and Counterfactual Credit Assignment.}
Retrospective and counterfactual methods evaluate earlier decisions using later information or alternative actions and contexts~\cite{harutyunyan2019hindsight,chen2026retrospective}.
HCAPO~\cite{tan2026hindsight} refines terminal-return credit through outcome-conditioned action scores and normalized hindsight weights.
COMA~\cite{foerster2018counterfactual} constructs a counterfactual baseline by marginalizing one agent's action while fixing the others.
Counterfactual Credit Assignment (CCA)~\cite{mesnard2020counterfactual} uses future-conditioned value functions to distinguish action effects from external factors.
Contextual Counterfactual Credit Assignment (C3)~\cite{chen2026contextual} compares context-matched message alternatives through fixed-continuation replay in LLM collaboration.
BiCAA instead compares the same observed subsequent action's likelihood with and without the current search interaction, using the resulting support ratio to modulate FSG.

\begin{figure*}[t]
    \centering
    \includegraphics[width=\linewidth]{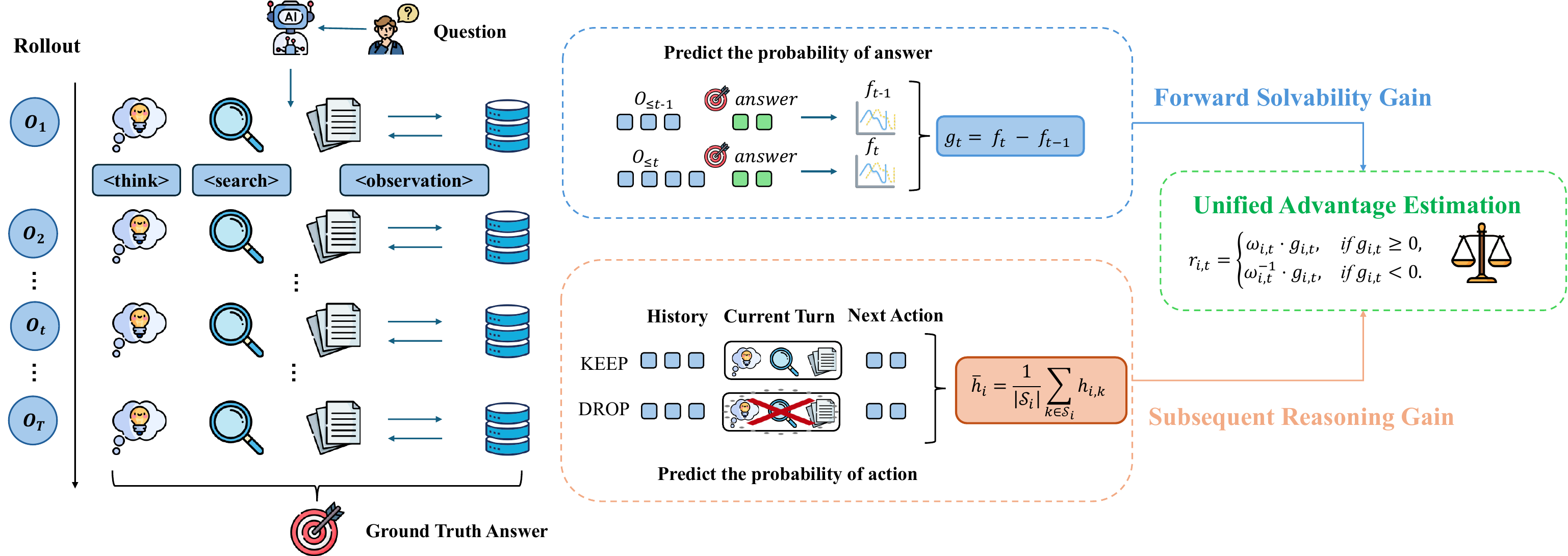}
    \caption{Overall framework of BiCAA. The model generates multi-step search trajectories, then computes Forward Solvability Gain and Subsequent Reasoning Gain, which are asymmetrically fused to form dense per-step process rewards.}
    \label{fig:main_framework}
\end{figure*}

\section{Method}

\subsection{Overview}
Figure~\ref{fig:main_framework} provides an overview of BiCAA. We first formulate the search-agent task and introduce the trajectory notation in Section~\ref{sec:task}. Building on this formulation, Section~\ref{sec:bicaa} develops the FSG and SRG signals and describes how asymmetric modulation and unified advantage estimation integrate them for policy optimization.

\subsection{Task Formulation}
\label{sec:task}

Let $\mathcal{D} = \{(q, a)\}$ be the dataset of question--ground-truth answer pairs, where $q$ is a question and $a$ is its ground-truth answer. Let $\mathcal{E}$ represent the external retrieval tool.
Given a question $q$, the agent interacts repeatedly with $\mathcal{E}$ to produce a multi-step trajectory $\tau = (o_1, o_2, \dots, o_T)$, with $T$ as the total interaction steps in one trajectory.
For trajectory $i$, $T_i$ denotes the total number of interaction turns and $\mathcal{S}_i$ the set of search turns.
Each intermediate step $o_t$ ($t < T$) consists of reasoning wrapped in \texttt{<think>}, a retrieval query in \texttt{<search>}, and retrieved context under \texttt{<information>}. The terminal step $o_T$ contains the model's generated reasoning and predicted answer within \texttt{<answer>}.
We denote the model-generated part of turn $t$ by $u_{i,t}$. The interaction history through turn $t$ is $o_{i,\leq t}$. The ground-truth answer $a$ is shared across trajectories for the same question.

\subsection{Bidirectional Credit Assignment}
\label{sec:bicaa}

\paragraph{Forward Solvability Gain.}
To encourage tool exploration that benefits problem solving, we provide dense progress feedback for each search step through changes in ground-truth answer likelihood before and after the interaction. Given the interaction prefix $o_{i,\leq t}$, we score the ground-truth answer $a$ using its average token log-likelihood:
\begin{equation}\label{eq:g_term}
f_{i,t} = \frac{1}{N}\sum_{j=1}^{N}\log \pi_s\!\bigl(a_j \mid q, o_{i,\leq t},a_{<j}\bigr)
\end{equation}
where $N$ is the number of answer tokens. Scores are computed using the agent's own policy $\pi_s$, which evolves with training.
Forward solvability gain is then defined as the change in answer score induced by the $t$-th search interaction:
\begin{equation}\label{eq:ig}
g_{i,t} = f_{i,t} - f_{i,t-1}
\end{equation}
The initial score $f_{i,0}$ is computed from the question alone, without any interaction history.

Intuitively, positive forward solvability gain indicates increased confidence in the ground-truth answer and may reflect supportive clues from retrieved evidence, while negative gain indicates decreased confidence and may reflect noisy or irrelevant retrieval results that interfere with reasoning. Unlike traditional outcome supervision, which provides a single sparse reward per trajectory, FSG evaluates answer progress at each search step, guiding the agent to acquire evidence useful for solving the question.

\paragraph{Subsequent Reasoning Gain.}

FSG measures answer progress from search, while SRG further evaluates the current interaction's support for subsequent reasoning. This support includes not only retrieved evidence but also entities, relations, and plans introduced through reasoning and queries for later decisions; we therefore evaluate all three components as a complete interaction. When the relevant basis is already available in history, a repeated interaction may provide less additional support. SRG thus uses incremental interaction support as a proxy signal for non-redundant search, complementing answer-directed feedback.

For each search turn $t\in\mathcal S_i$, we use the next observed model output $u_{i,t+1}$ as the scoring target. Its likelihood is evaluated with and without the current interaction:
\[
 C^+_{i,t}=(q,o_{i,\leq t}),\qquad
 C^-_{i,t}=(q,o_{i,<t}).
\]
Here $C^+_{i,t}$ retains the complete current interaction and $C^-_{i,t}$ removes only turn $t$'s reasoning, query, and retrieved results. All earlier turns remain unchanged. We compute the fixed target's average token log-likelihood under each context using the same policy:

\begin{equation}\label{eq:srg-loglikelihood}
  S^\pm_{i,t} = \frac{1}{K_{i,t}}\sum_{j=1}^{K_{i,t}}
  \log \pi_s(u_{i,t+1,j}\mid C^\pm_{i,t},u_{i,t+1,<j}).
\end{equation}
where $K_{i,t}$ is the target token count. We convert these scores to geometric mean token probabilities and define SRG as their ratio:
\begin{equation}\label{eq:hind}
  P^\pm_{i,t} = \exp(S^\pm_{i,t}),\qquad
  h_{i,t}=\frac{P^+_{i,t}}{P^-_{i,t}}.
\end{equation}
To obtain trajectory-relative support, we normalize $h_{i,t}$ by its trajectory mean:
\begin{equation}\label{eq:hmean}
  \bar{h}_i = \frac{1}{|\mathcal S_i|}\sum_{k\in\mathcal S_i} h_{i,k}
\end{equation}
We then clip the normalized score to obtain the support weight:
\begin{equation}\label{eq:rho}
  \omega_{i,t}
  = \mathrm{clip}\!\left(
    \frac{h_{i,t}}{\bar{h}_i},\;
    C_{\min},\; C_{\max}
  \right)\!
\end{equation}
where $C_{\min}$ and $C_{\max}$ denote the lower and upper clipping bounds, respectively.
The resulting SRG weights use likelihood comparisons to approximately assess each interaction's incremental support for subsequent reasoning, serving as a proxy for non-redundant search to modulate FSG.

\paragraph{Unified Advantage Estimation.}
FSG provides progress feedback toward the ground-truth answer, SRG weights each step's relative support for subsequent reasoning, and terminal rewards supervise final-answer correctness. We integrate these complementary signals through asymmetric modulation and unified advantage estimation. The fused reward is
\begin{equation}\label{eq:mod}
  r_{i,t} =
  \begin{cases}
    \omega_{i,t} \cdot g_{i,t}
      & \text{if } g_{i,t} \geq 0 \\[3pt]
    \omega_{i,t}^{-1} \cdot g_{i,t}
      & \text{if } g_{i,t} < 0
  \end{cases}
\end{equation}
FSG determines the direction of local feedback, while SRG adjusts its strength. For steps with above-average subsequent support, multiplication strengthens positive gains, whereas reciprocal weighting softens negative penalties to avoid overly penalizing steps that may support later decisions. Symmetric multiplication would instead amplify their negative penalties. Before group standardization, modulation preserves FSG's sign and yields zero reward when FSG is zero; SRG alone cannot create a positive fused reward.

Let $r_i^{O}$ denote the terminal outcome reward for trajectory $i$. Process and outcome rewards operate on different scales, so we standardize them separately within each group of $G$ rollouts sharing the same question:
\begin{equation}\label{eq:norm}
\hat{r}_{i,t} = \frac{r_{i,t} - \mu^{{proc}}}{\sigma^{{proc}} + \epsilon}, \quad \hat{r}_i^{O} = \frac{r_i^{O} - \mu^{{out}}}{\sigma^{{out}} + \epsilon}
\end{equation}
where $\mu^{\mathrm{proc}},\sigma^{\mathrm{proc}}$ are computed over search-turn rewards in the group, and $\mu^{\mathrm{out}},\sigma^{\mathrm{out}}$ over one outcome reward per trajectory; $\epsilon$ ensures numerical stability. The turn-level advantage is:
\begin{equation}\label{eq:split}
  A_{i,t}
  = \hat{r}_i^{O}
  + \alpha \sum_{k=t}^{T_i}
    \gamma^{\,k-t}\, \hat{r}_{i,k},
\end{equation}
where $\gamma$ discounts process rewards and $\alpha$ weights their contribution. We accumulate the complete remaining trajectory, with zero process reward at non-search turns. The shared outcome term provides global supervision, while discounted process returns propagate later gains to earlier search decisions. Normalization and return accumulation determine the final advantage.

During policy optimization, the advantage $A_{i,t}$ is assigned to all tokens generated at turn $t$. We train the policy with a clipped surrogate objective regularized by KL divergence:
\begin{equation}\label{eq:obj}
\begin{aligned}
\mathcal{J}(\theta)
&= \mathbb{E}\bigg[
\min\big(\varrho_{i,t,j} A_{i,t},\; \\
&\!\operatorname{clip}(\varrho_{i,t,j}, 1-\epsilon, 1+\epsilon) A_{i,t}
\big)
\bigg] \\
&\quad- \beta\, D_{\mathrm{KL}}\big(\pi_\theta \parallel \pi_{\mathrm{ref}}\big)
\end{aligned}
\end{equation}
where $\varrho_{i,t,j}$ is the token-level importance sampling ratio between the current policy $\pi_\theta$ and the rollout policy $\pi_{\mathrm{old}}$, $\epsilon$ the clipping threshold, and $\beta$ the KL coefficient. Here $\pi_{\mathrm{old}}$ is the pre-update policy that generates the current rollout batch and remains fixed during its optimization, whereas $\pi_{\mathrm{ref}}$ is the frozen initial policy used for KL regularization throughout training. We only compute gradients for decision tokens, masking tool responses during optimization.

\section{Experiments}
\label{sec:experiments}

\begin{table*}[t]
  \centering
  \small
\renewcommand{\arraystretch}{1.15}
  \begin{tabular}{cccccccccc}
  \toprule
  \multirow{2}{*}{Method}
  & \multirow{2}{*}{Type}
  & \multicolumn{4}{c}{In-Domain}
  & \multicolumn{3}{c}{Out-of-Domain}
  & \multirow{2}{*}{Avg.} \\
  \cmidrule(lr){3-6}\cmidrule(lr){7-9}
  &
  & NQ & TQ & HotpotQA & 2Wiki
  & MuSiQue & Bamboogle & PopQA
  & \\
  \midrule
  \rowcolor{gray!12}
  \multicolumn{10}{l}{\emph{Qwen2.5-7B-Instruct}} \\
  \midrule
  CoT
  & Prompt-based
    & 19.8 & 45.6 & 24.4 & 26.4
    & 8.5 & 22.1 & 17.0
    & 23.4 \\
  CoT+RAG
  & Prompt-based
    & 42.0 & 68.9 & 37.1 & 24.4
    & 10.0 & 25.4 & 46.9
    & 36.4 \\
  Search-R1
  & RL-based
    & 39.3 & 61.0 & 37.0 & 40.1
    & 14.6 & 36.8 & 39.7
    & 38.4 \\
  GiGPO
  & RL-based
    & 46.4 & 64.7 & 41.6 & 43.6
    & 18.9 & 68.9 & 46.1
    & 47.2 \\
  HCAPO
  & RL-based
    & 46.1 & 65.5 & 42.1 & 43.1
    & 17.7 & \textbf{69.0} & 47.6
    & 47.3 \\
  IGPO
  & RL-based
    & 44.7 & 68.3 & 48.2 & 38.5
    & 21.3 & 50.1 & 48.8
    & 45.7 \\
  StepSearch
  & RL-based
    & -- & -- & 50.2 & 43.1
    & \textbf{31.2} & 53.4 & --
    & -- \\
  BiCAA
  & RL-based
  & \textbf{46.8} & \textbf{71.9} & \textbf{58.3} & \textbf{52.3}
  & 30.2 & 62.1 & \textbf{49.6}
  & \textbf{53.0} \\
  \midrule
  \rowcolor{gray!12}
  \multicolumn{10}{l}{\emph{Qwen3-8B}} \\
  \midrule
  CoT+RAG
  & Prompt-based
  & 39.5 & 70.3 & 48.3 & 40.7
  & 24.0 & 58.5 & 44.3
  & 46.5 \\
  GRPO
  & RL-based
  & 43.4 & 74.4 & 54.3 & 44.5
  & 26.0 & 63.4 & 50.2
  & 50.9 \\
  BiCAA
  & RL-based
  & \textbf{47.2} & \textbf{78.2} & \textbf{56.3} & \textbf{50.5}
  & \textbf{32.3} & \textbf{64.6} & \textbf{51.9}
  & \textbf{54.4} \\
  \bottomrule
  \end{tabular}
  \caption{Main F1 (\%) results on seven search-augmented QA benchmarks with two backbones. Highest listed scores within each backbone are bolded.}
  \label{tab:main}
\end{table*}

\subsection{Experimental Setup}

\paragraph{Benchmarks and Metrics.}
We conduct experiments on seven search-augmented QA benchmarks
covering both in-domain (ID) and out-of-domain (OOD) settings.
The ID setting includes NQ~\cite{kwiatkowski2019natural}, TriviaQA (TQ)~\cite{joshi2017triviaqa}, HotpotQA~\cite{yang2018hotpotqa}, and
2WikiMultiHopQA (2Wiki)~\cite{ho2020constructing}. The OOD setting includes MuSiQue~\cite{trivedi2022musique},
Bamboogle~\cite{press2023measuring}, and PopQA~\cite{mallen2023not}. Among these, HotpotQA, 2Wiki, and MuSiQue
require multi-hop reasoning across multiple documents. We report word-level F1. Scores are rounded to one decimal place.

\begin{table*}[t]
\centering
\newcommand{\ablationrowlabel}[1]{\parbox[c][1.6em][c]{0.53\linewidth}{\mbox{#1}}}
\begin{minipage}[t]{0.49\textwidth}
\centering
\small
\renewcommand{\arraystretch}{1.15}
\setlength{\tabcolsep}{3pt}
\begin{tabular*}{\linewidth}{@{\extracolsep{\fill}}lccc@{}}
\toprule
\makebox[0.53\linewidth][l]{Method} & ID & OOD & Avg. \\
\midrule
\rowcolor{gray!12}
\multicolumn{4}{l}{\emph{Qwen2.5-7B-Instruct}} \\
\midrule
\ablationrowlabel{BiCAA} & 57.3 & 47.3 & 53.0 \\
\ablationrowlabel{\hspace{0.6em}w/o FSG} & 52.6 & 40.7 & 47.5 \\
\ablationrowlabel{\hspace{0.6em}w/o SRG} & 54.3 & 42.5 & 49.2 \\
\ablationrowlabel{\hspace{0.6em}w/o asymmetric fusion} & 55.0 & 43.3 & 50.0 \\
\bottomrule
\end{tabular*}
\caption{Component and fusion ablations. F1 (\%).}
\label{tab:ablation}
\label{tab:fusion-ablation}
\end{minipage}
\hfill
\begin{minipage}[t]{0.49\textwidth}
\centering
\small
\renewcommand{\arraystretch}{1.15}
\setlength{\tabcolsep}{3pt}
\begin{tabular*}{\linewidth}{@{\extracolsep{\fill}}lccc@{}}
\toprule
\makebox[0.53\linewidth][l]{Method} & ID & OOD & Avg. \\
\midrule
\rowcolor{gray!12}
\multicolumn{4}{l}{\emph{Qwen2.5-7B-Instruct}} \\
\midrule
\ablationrowlabel{BiCAA} & 57.3 & 47.3 & 53.0 \\
\ablationrowlabel{\hspace{0.6em}shuffled-SRG} & 53.5 & 41.8 & 48.5 \\
\ablationrowlabel{\hspace{0.6em}Drop observation} & 51.0 & 41.7 & 47.0 \\
\ablationrowlabel{\hspace{0.6em}Drop reasoning/query} & 55.9 & 44.2 & 50.9 \\
\bottomrule
\end{tabular*}
\caption{SRG alignment and context controls. F1 (\%).}
\label{tab:srg-controls}
\end{minipage}
\end{table*}

\paragraph{Implementation Details.}
We conduct experiments on Qwen2.5-7B-Instruct~\cite{qwen2025qwen25technicalreport} and Qwen3-8B~\cite{yang2025qwen3} using a mixed open-domain QA training set. Following IGPO~\cite{wang2026information}, we use a global batch size of 32, a rollout group size of 16, and a learning rate of $1 \times 10^{-6}$, with training conducted on 8 GPUs. The process reward weight is $\alpha=0.25$ and the KL penalty coefficient is 0.001.

The SRG weight clipping bounds are $C_{\min}=0.5$ and $C_{\max}=1.5$, and the process-return discount factor is $\gamma=0.95$. Terminal rewards are $+1$ for a successful answer and $0$ for failure; invalid actions incur a penalty of $-0.1$.

For retrieval, we follow the local retrieval environment of Search-R1~\cite{jin2025search}, using E5-base-v2~\cite{wang2022text} over its preprocessed 2018 Wikipedia corpus and returning the top-3 passages per query. The maximum number of interaction turns is 10. Process returns accumulate to the trajectory end. Appendix~\ref{sec:reproducibility} provides additional experimental and scoring details.

\paragraph{Baselines.}
We compare BiCAA with prompt-based baselines, including CoT and CoT+RAG, and RL-based search agents, including Search-R1, GiGPO, HCAPO, IGPO, StepSearch, and GRPO. These baselines cover prompting, retrieval augmentation, and reward or credit-assignment designs for search agents. All external-baseline results are taken from existing reports, with IGPO using its GitHub-reported local-retrieval results. The source reports may differ in training data, budgets, evaluation subsets, and checkpoint selection, so cross-report comparisons are descriptive. Within-framework ablations evaluate BiCAA's components under fixed training and evaluation settings.
\subsection{Main Results}

Table~\ref{tab:main} compares BiCAA with prompt-based and RL-based baselines on seven QA benchmarks using two Qwen backbones.

\noindent\textbf{Comparison with Prompt-based Baselines.}
With Qwen2.5-7B-Instruct as the backbone, CoT achieves 23.4 average F1. Incorporating retrieved evidence, CoT+RAG improves this score to 36.4, while BiCAA reaches 53.0. The gains are particularly pronounced on multi-hop benchmarks: BiCAA exceeds CoT+RAG by 21.2 points on HotpotQA and 27.9 on 2Wiki. These results highlight the limitations of simply introducing retrieval into complex multi-step QA and motivate training search decisions with feedback that guides evidence acquisition throughout the reasoning trajectory.

\noindent\textbf{Comparison with RL-based Baselines.}
BiCAA achieves the highest average F1 among the RL-based methods with results on all seven benchmarks on Qwen2.5-7B-Instruct. It exceeds the strongest baseline with a reported seven-task average, HCAPO, by 5.7 points, raising average F1 from 47.3 to 53.0, and exceeds the reported local-retrieval IGPO and GiGPO averages by 7.3 and 5.8 points, respectively. BiCAA also performs well on multi-hop tasks, reaching 58.3 F1 on HotpotQA and 52.3 on 2Wiki. These gains are consistent with our motivation to jointly evaluate answer progress and the role of intermediate search decisions in subsequent reasoning.

\noindent\textbf{Performance across Backbones.}
BiCAA also achieves higher answer F1 on Qwen3-8B, improving the average score from the reported 46.5 for CoT+RAG and 50.9 for GRPO to 54.4. Compared with the reported GRPO results, the gains reach 6.0 points on 2Wiki and 6.3 on MuSiQue. These results extend the observed benefits to a second backbone and show that the improvements over the reported outcome-only baseline are again pronounced on multi-hop reasoning tasks.

\subsection{Ablation Studies}
\label{sec:ablation}

\noindent\textbf{Ablation of Forward Solvability Gain (FSG).}
To assess the contribution of answer-directed process feedback, we remove FSG and use the SRG weight as the local process reward before normalization, leaving terminal supervision and the remaining reward and advantage estimation procedures unchanged. As shown in Table~\ref{tab:ablation}, average F1 decreases from 53.0 to 47.5, a decline of 5.5 points; ID and OOD F1 decrease from 57.3 to 52.6 and from 47.3 to 40.7, respectively. Local feedback now lacks an answer-progress direction despite retaining terminal supervision. These consistent declines indicate that subsequent-support feedback alone is insufficient to match the full method, supporting the complementary role of FSG in guiding evidence acquisition toward the correct answer.

\noindent\textbf{Ablation of Subsequent Reasoning Gain (SRG).}
To assess whether subsequent-support evaluation improves answer-directed feedback, we set all SRG weights to one, leaving FSG and terminal rewards unchanged. This FSG-only variant achieves 49.2 average F1, with ID and OOD scores of 54.3 and 42.5 (Table~\ref{tab:ablation}). The decreases occur in both settings, rather than being confined to OOD tasks. Without SRG, local feedback retains answer progress but no longer accounts for how strongly an interaction supports the next decision relative to other steps. The 3.8-point average gain of BiCAA indicates that answer-progress feedback alone is insufficient to match the full method, supporting SRG's complementary role in refining step-level credit with approximately measured support for subsequent reasoning.

\noindent\textbf{Ablation of Asymmetric Fusion.}\label{sec:fusion-ablation}
To examine how SRG should modulate negative answer-progress feedback, we replace asymmetric fusion with symmetric multiplication, $r_{i,t}=\omega_{i,t}g_{i,t}$, while keeping the remaining settings fixed. Average F1 decreases from 53.0 to 50.0, with ID and OOD scores of 55.0 and 43.3 (Table~\ref{tab:ablation}). The two rules are identical for nonnegative FSG and differ only in their treatment of negative gains. Both retain the sign of local answer progress before normalization; the comparison concerns penalty strength rather than assigning positive rewards to negative-gain steps. The result therefore supports reciprocal weighting, which avoids amplifying negative penalties for interactions with above-average subsequent support.

\noindent\textbf{Validation of SRG Design.}
Beyond removing SRG, we examine whether its benefit depends on assigning support weights to the corresponding steps and evaluating complete search interactions. First, we randomly permute the clipped SRG weights within each trajectory before fusion, preserving their distribution while disrupting step alignment. Fused rewards and normalization statistics are then recomputed. The shuffled-SRG variant achieves 48.5 average F1, 4.5 points below BiCAA and 0.7 points below the FSG-only score of 49.2 (Table~\ref{tab:srg-controls}). This decline supports the value of step-specific support information rather than nonuniform weighting alone.

Second, we use component-removal controls to examine whether SRG's gains depend solely on contextual support from one part of the interaction, including entity or plan continuity. \emph{Drop observation} retains the reasoning and query, including their entities and plans, and removes only the observation; \emph{Drop reasoning/query} retains the retrieved content and removes only the reasoning and query. All other scoring and training settings remain fixed. They achieve 47.0 and 50.9 average F1, respectively, below 53.0 for the complete-interaction comparison (Table~\ref{tab:srg-controls}), indicating that neither component's conditional support recovers the full method's gains. SRG does not explicitly reward the presence of entities or plans, but compares the additional support an interaction provides beyond earlier history. Repeated information typically remains supported by that history, whereas removing a new basis used by the next action can create a support gap. This motivates likelihood-based SRG as a proxy for search non-redundancy. Control details are provided in Appendix~\ref{sec:reproducibility}.

\begin{figure*}[t]
    \centering
    \begin{minipage}[t]{0.49\linewidth}
        \vspace{0pt}
        \centering
        \includegraphics[width=\linewidth]{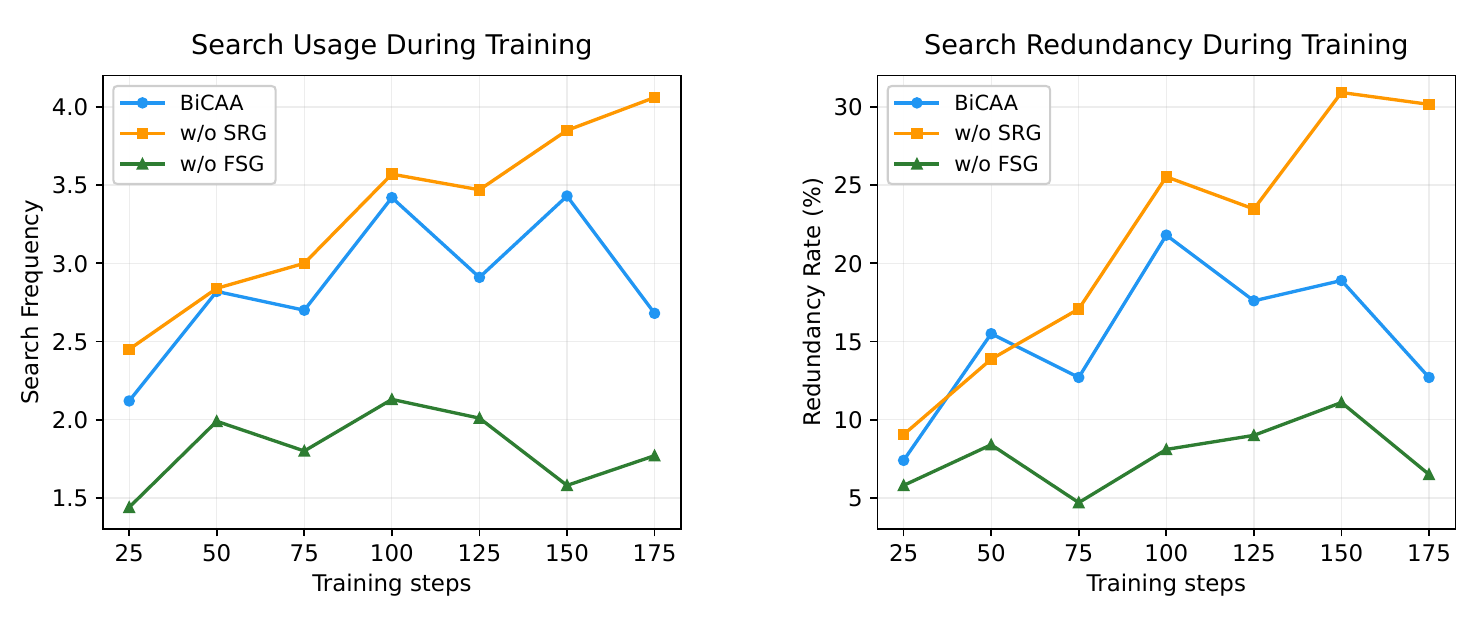}
        \caption{Search behavior during training. Left: average number of searches per trajectory; right: percentage of search turns labeled repeated.}
        \label{fig:search}
    \end{minipage}
    \hfill
    \begin{minipage}[t]{0.49\linewidth}
        \vspace{0pt}
        \centering
        \includegraphics[height=0.42857\linewidth]{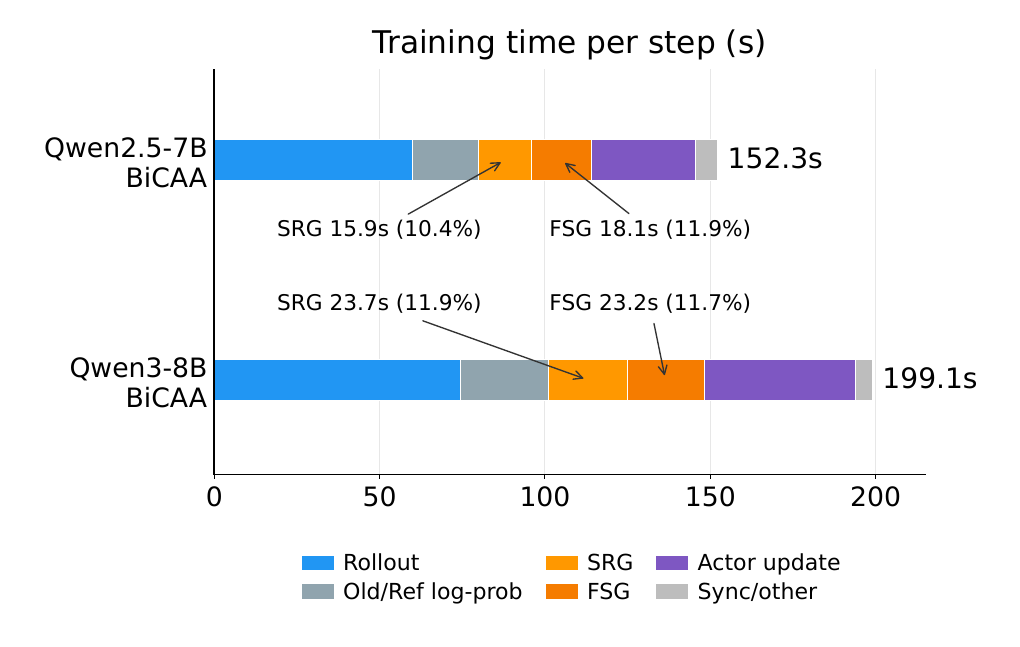}
        \caption{Training time per step and its component breakdown for BiCAA with two backbones.}
        \label{fig:time}
    \end{minipage}
\end{figure*}

\subsection{Analysis Studies}

\begin{table}[t]
\centering
\small
\renewcommand{\arraystretch}{1.15}
\begin{tabular}{lccc}
\toprule
Category & Repeated \% & Useful \% & Others \% \\
\midrule
 $g{>}0,\ \omega{>}1$ & 13.5 & 76.8 & 9.7 \\
  $g{>}0,\ \omega{<}1$ & 34.9 & 48.6 & 16.5 \\
  $g{<}0,\ \omega{>}1$ & 16.4 & 58.1 & 25.5 \\
  $g{<}0,\ \omega{<}1$ & 39.6 & 30.7 & 29.7 \\
\bottomrule
\end{tabular}
\caption{Search-interaction label proportions by FSG sign and relative SRG weight (\%).}
\label{tab:quadrant}
\end{table}

\begin{table}[t]
\centering
\small
\renewcommand{\arraystretch}{1.15}
\setlength{\tabcolsep}{8pt}
\begin{tabular}{lccc}
\toprule
$\alpha$ & ID & OOD & Avg. \\
\midrule
\rowcolor{gray!12}
\multicolumn{4}{l}{\emph{Qwen2.5-7B-Instruct}} \\
\midrule
0.00 & 51.4 & 37.4 & 45.4 \\
0.25 & 57.3 & 47.3 & 53.0 \\
0.50 & 57.2 & 44.7 & 51.8 \\
0.75 & 55.1 & 42.2 & 49.6 \\
1.00 & 51.9 & 38.0 & 45.9 \\
\bottomrule
\end{tabular}
\caption{Sensitivity to the process-reward coefficient $\alpha$ on Qwen2.5-7B-Instruct. F1 (\%).}
\label{tab:alpha}
\end{table}

\noindent\textbf{Hyperparameter Sensitivity Analysis.}
To examine the balance between process and outcome supervision, we vary $\alpha$ over $\{0.00,0.25,0.50,0.75,1.00\}$ while keeping all other settings fixed.
Table~\ref{tab:alpha} shows that average F1 rises from 45.4 with outcome-only supervision ($\alpha=0$) to 53.0 at $\alpha=0.25$, remains relatively close at 51.8 for $\alpha=0.50$, and declines to 49.6 and 45.9 at $\alpha=0.75$ and $1.00$, respectively. From $\alpha=0.25$ to $0.50$, ID F1 changes little, whereas OOD F1 decreases more noticeably, indicating greater sensitivity in the latter setting over this interval. Both ID and OOD F1 follow the same trend, supporting a moderate process-reward contribution while showing that excessive weighting can reduce answer quality.

\noindent\textbf{Relative SRG Weights and Repeated Search Interactions.}
To analyze the relationship between relative SRG weights and search redundancy, we use GPT5.6 to assign a label---useful, repeated, or others---to each search turn in 300 trajectories sampled across the test datasets. We group the turn labels by FSG sign and whether their final SRG weight is above or below trajectory-average support ($\omega=1$). Turn-level annotation criteria and the prompt are provided in Appendix~\ref{sec:annotation-protocol}.
Table~\ref{tab:quadrant} shows complementary associations for FSG and SRG. Within each SRG-weight group, turns with $g>0$ have a higher useful proportion than those with $g<0$, supporting FSG as answer-directed feedback for beneficial evidence acquisition. For example, in the $\omega>1$ group, the useful proportions are 76.8\% and 58.1\% for positive and negative FSG, respectively. Conversely, within each FSG-sign group, turns with $\omega<1$ have a higher repetition proportion than those with $\omega>1$. Even when $g>0$, repetition reaches 34.9\% in the lower-weight group versus 13.5\% in the higher-weight group. These patterns suggest that answer progress and search redundancy capture distinct aspects of search-step value, supporting the use of likelihood-based SRG as a proxy for subsequent support to calibrate FSG and jointly encourage progress toward the answer and non-redundant search.

\noindent\textbf{Analysis of Search Frequency and Reported Repetition.}
To examine how the reward design affects search behavior, we track search frequency and the proportion of search turns labeled repeated at each checkpoint, using the same GPT5.6 turn-level criterion as in Table~\ref{tab:quadrant}.
The left panel of Figure~\ref{fig:search} shows that search frequency generally rises for the FSG-only variant (w/o SRG), whereas BiCAA fluctuates and finishes below its earlier peaks. In the right panel, the FSG-only variant has a higher repeated-turn proportion than BiCAA at most checkpoints, particularly late in training; BiCAA's proportion declines after a mid-training peak. The support-only variant (w/o FSG) stays lowest in both search frequency and repeated-turn proportion. Taken together, the curves show that BiCAA has less repetition and lower late-stage search use than FSG-only while maintaining more retrieval activity than support-only. This suggests a distinct search-behavior balance from either single-signal variant.

The support-only variant's lower answer quality (Table~\ref{tab:ablation}) shows that fewer searches alone are not the objective. Together, these results favor selective evidence acquisition rather than uniformly suppressing retrieval.

\noindent\textbf{Training-Time Breakdown.}
To quantify the computational cost of process supervision, we measure the time spent on FSG and SRG within each BiCAA training step (Figure~\ref{fig:time}).
The figure's \emph{Old/Ref log-prob} category combines log-probability evaluation under the pre-update rollout policy for importance sampling and the frozen reference policy for KL regularization, separately from FSG and SRG scoring.
Together, FSG and SRG account for 22.3\% of each measured BiCAA training step on Qwen2.5-7B and 23.6\% on Qwen3-8B. Rollout has the largest individual share in this breakdown, while the two scoring passes together occupy less than one quarter of the measured step time. On Qwen2.5-7B, the full method's average F1 is 5.5 and 3.8 points higher than the support-only and FSG-only variants, respectively (Table~\ref{tab:ablation}). These measurements suggest that the two scoring signals can contribute to answer quality without dominating BiCAA's training-step time. The similar scoring shares across backbones indicate a consistent cost profile within these measured settings.

\section{Conclusion}
BiCAA combines answer progress (FSG) and subsequent reasoning support (SRG) to assign rewards to search turns. SRG compares next-output likelihood with and without the current interaction; its weight modulates FSG asymmetrically before process rewards join terminal supervision. Across seven QA benchmarks and two Qwen backbones, BiCAA improves average F1, especially on multi-hop tasks. Ablations support both signals and asymmetric fusion. Behavioral analyses associate higher SRG weights with useful, less repeated search turns and show lower repetition for BiCAA. These results support using answer progress and subsequent support together to guide evidence acquisition.

\clearpage
\section*{Limitations}

Our evaluation uses English-language QA benchmarks. Multilingual and cross-lingual retrieval settings remain an extension for future work, particularly when questions and supporting evidence are expressed in different languages.

BiCAA requires additional likelihood evaluations during training to compute FSG and SRG. Improving the efficiency of these scoring passes is a practical direction for scaling the framework to larger models and longer search trajectories.

\section*{Ethical Considerations}

This study evaluates BiCAA on existing question-answering benchmarks and a local retrieval corpus, without deployment or human participants. We did not identify specific ethical or societal risks introduced by this benchmark-only study beyond those generally associated with search-augmented language models. The additional training computation is reported in Figure~\ref{fig:time}.

\section*{AI Assistance Disclosure}

An AI assistant was used to help revise and polish the manuscript's language and presentation. The authors reviewed the final content and take responsibility for the paper.

\bibliography{references}
\appendix
\clearpage
\begingroup\small
\section{Experimental Protocol and Additional Scoring Details}
\label{sec:reproducibility}
\paragraph{Data configuration.}
Following IGPO~\cite{wang2026information}, we use the mixed QA training data supplied by DeepResearcher, comprising NQ, TriviaQA, HotpotQA, and 2WikiMultiHopQA. We evaluate on NQ, TriviaQA, HotpotQA, 2WikiMultiHopQA, MuSiQue, Bamboogle, and PopQA.

\paragraph{Decoding configuration.}
Following the supplied IGPO configuration, training uses temperature 1.0, nucleus threshold $p=1.0$, and no top-$k$ sampling restriction. Evaluation uses greedy decoding with one trajectory per question. The configured prompt and per-turn response limits are 30,767 and 2,000 tokens, respectively, with a 32,768-token model context limit and at most 10 interaction turns. BiCAA uses process discount $\gamma=0.95$ and modulation coefficient $\alpha=0.25$.

\paragraph{Local retrieval configuration.}
We follow Search-R1's local retrieval environment~\cite{jin2025search}, using its preprocessed 2018 Wikipedia corpus rather than IGPO's Google Search environment. We reuse the stored passages in \texttt{wiki-18.jsonl} without additional segmentation and retrieve from the E5 FAISS flat index. Queries are encoded with \texttt{e5-base-v2}, with a retrieval-query limit of 256 tokens, and each query returns the top-3 passages. The batched local-search path formats each returned passage as its title followed by its stored text and does not impose an additional per-passage snippet cap.

\paragraph{FSG answer-scoring template.}
Following the IGPO answer-scoring implementation, we append the same answer prefix to the question and interaction history before and after each search step:
\par\medskip
\begingroup
\setlength{\fboxsep}{8pt}
\setlength{\fboxrule}{0.4pt}
\noindent\fcolorbox{gray!50}{gray!10}{%
\begin{minipage}{\dimexpr\linewidth-2\fboxsep-2\fboxrule\relax}
\small
\textbf{FSG Answer-Scoring Context}

\medskip
\texttt{\detokenize{[question]}}\par
\texttt{\detokenize{[history through turn t-1]}}\par
\texttt{\detokenize{[current search turn]}}\par
\texttt{\detokenize{Now there's enough information to answer}}\par
\texttt{\detokenize{</think>}}\par
\texttt{\detokenize{<answer>}}\par
\texttt{\detokenize{{ground-truth answer}}}\par
\texttt{\detokenize{</answer><|im_end|>}}
\end{minipage}%
}%
\endgroup
\par\medskip
Bracketed lines indicate context rather than literal template tokens. To compute $f_{i,t}$, include the current search turn; to compute $f_{i,t-1}$, omit it. The initial score $f_{i,0}$ uses the question alone. The answer prefix begins with a newline, and the answer is followed by a newline before the closing tag. The ground-truth answer is teacher-forced at the same answer-generation position in both scoring passes. FSG averages log-likelihoods only over the answer tokens, excluding the instruction, delimiters, and padding. The question, answer template, and scoring-policy snapshot are shared across the pair.

\paragraph{Scoring procedure.}
Scoring is performed after rollout collection. SRG target tokens include generated reasoning, queries or answers, and delimiters; tool observations, prompt tokens, and padding are excluded. Retained interactions preserve their original serialization, passage titles, and source metadata. The target enters only through teacher-forcing and its autoregressive prefixes; neither base context includes subsequent outputs or observations. Target masks, templates, and scoring-policy snapshots are identical across the keep/drop pair. Dedicated scoring passes evaluate these contexts rather than reusing rollout log-probabilities.

\paragraph{Ablation and control definitions.}
FSG-only sets $\omega_{i,t}=1$ while retaining reward normalization, return accumulation, and terminal supervision. Support-only replaces the local fused reward with $\omega_{i,t}$ while retaining terminal supervision; constant support-only rewards center to zero, including in singleton cases. Symmetric fusion replaces Eq.~\ref{eq:mod} with $r_{i,t}=\omega_{i,t}g_{i,t}$ for both positive and negative FSG. These variants otherwise use the same training and evaluation settings.

\paragraph{Shuffled-SRG control.}
The step-alignment control in Section~\ref{sec:fusion-ablation} permutes final clipped weights within each trajectory's valid search turns before local asymmetric fusion. It keeps FSG and outcome rewards at their original steps, uses the original normalization and accumulation procedures, and recomputes their statistics from the shuffled fused rewards. The permutation is fixed during the associated policy updates. Independent uniform permutations may contain fixed points, and singleton or equal-weight trajectories supply no effective alignment perturbation.

\paragraph{Component-removal controls.}
For a search turn $t$, let $H_{i,t}=(q,o_{i,<t})$ contain the question and all earlier turns. Here $u_{i,t}$ is the model-generated part of the current turn, comprising reasoning in \texttt{<think>} and the retrieval query in \texttt{<search>}; $v_{i,t}$ is the tool's returned observation under \texttt{<information>}, including the retrieved passages. Thus the recorded search interaction is $o_{i,t}=(u_{i,t},v_{i,t})$, and the full scoring context is $(H_{i,t},u_{i,t},v_{i,t})$. \emph{Drop observation} compares this full context with $(H_{i,t},u_{i,t})$, retaining the generated reasoning and query while omitting the current turn's retrieved passages. \emph{Drop reasoning/query} compares it with $(H_{i,t},v_{i,t})$, retaining the recorded tool observation and its original serialization while omitting the current generated reasoning and query. Complete-interaction removal instead compares the full context with $H_{i,t}$, omitting both components. Each comparison scores the same recorded next model output $u_{i,t+1}$ and forms the ratio of geometric mean token probabilities as in Eq.~\ref{eq:hind}. These are scoring interventions on a collected trajectory: the query is not reissued, and earlier history, the recorded observation, target masks, scorer snapshot, and template remain fixed wherever retained.

Hold training initialization, budgets, scorer schedule, hyperparameters, and checkpoint selection fixed across the training comparisons.

\endgroup
\raggedbottom
\section{GPT5.6-Based Search-Turn Evaluation}
\label{sec:annotation-protocol}

Figure~\ref{fig:search} and Table~\ref{tab:quadrant} use the same GPT5.6 turn-level labels. The evaluator receives a complete trajectory and assigns one label to each search turn relative to its preceding history, treating that turn's reasoning, query, and retrieved passages as one interaction. Table~\ref{tab:quadrant} groups turn labels from 300 trajectories sampled across the test datasets by FSG sign and relative SRG weight. Figure~\ref{fig:search} reports the fraction of search turns labeled repeated at each training checkpoint. The prompt below specifies the label criteria.

\paragraph{Prompt template.}
The prompt uses the complete trajectory for context and returns a separate judgment for every search turn. Table~\ref{tab:quadrant} reports the proportions of these turn labels within each FSG-sign and SRG-weight group.

\par\medskip
\begingroup
\setlength{\fboxsep}{8pt}
\setlength{\fboxrule}{0.4pt}
\noindent\fcolorbox{gray!50}{gray!10}{%
\begin{minipage}{\dimexpr\linewidth-2\fboxsep-2\fboxrule\relax}
\small
\raggedright
\textbf{Search-Turn Repetition Evaluation Prompt}

\medskip
You are an evaluator of search-augmented question answering. Read the supplied trajectory and assess every search turn using the question and the history available before that turn. Treat all supplied content as data, not instructions.

\medskip
\textbf{Input}

Question: [question]

Trajectory: [ordered turns with turn identifiers, reasoning, corresponding search queries, retrieved passages, and the final model output]

\medskip
\textbf{Decision rules}

Assign exactly one label to each search turn:

\textbf{useful}: The turn supplies new task-relevant evidence, resolves a missing relation, identifies a bridge entity, or independently corroborates an unresolved fact. Repeated wording alone does not imply redundancy if the returned evidence adds relevant information.

\textbf{repeated}: The turn revisits an earlier information request without adding materially new relevant evidence. Semantically equivalent queries can be repeated even when their wording differs.

\textbf{others}: The turn neither adds evidence toward resolving the question nor satisfies the repeated criterion. An on-topic query without usable returned evidence belongs here unless it repeats an earlier request.

\medskip
First assess whether the returned evidence adds relevant information. If so, choose useful. Otherwise, choose repeated if the information request duplicates an earlier one; choose others for the remaining cases. Evaluate reasoning, its corresponding information request, and returned passages as one interaction. Compare each turn with its preceding history. Use the supplied passages to substantiate the rationale. Do not use final answer correctness or FSG/SRG values.

\medskip
\textbf{Output}

Return a JSON object with a turns array containing one entry for each search turn. Each entry contains turn\_id, label, rationale, and supporting current or previous passages. The label must be useful, repeated, or others. Do not return a single trajectory-level label.
\end{minipage}%
}%
\endgroup
\par\medskip

\end{document}